\documentclass[runningheads]{llncs}
\usepackage{graphicx}
\usepackage{color, colortbl}
\usepackage{booktabs} 
\usepackage{makecell, multirow}
\usepackage{amsmath,amsfonts}
\usepackage{hyperref}
\usepackage{graphbox}
\definecolor{lightgray}{rgb}{.9,.9,.9}
\DeclareMathOperator{\transencoder}{TransformerEncoder}
\DeclareMathOperator{\softmax}{Softmax}

\usepackage{tikz}

\definecolor{lime}{HTML}{A6CE39}
\DeclareRobustCommand{\orcidicon}{%
	\begin{tikzpicture}
	\draw[lime, fill=lime] (0,0) 
	circle [radius=0.16] 
	node[white] {{\fontfamily{qag}\selectfont \tiny ID}};
	\draw[white, fill=white] (-0.0625,0.095) 
	circle [radius=0.007];
	\end{tikzpicture}
	\hspace{-2mm}
}

\foreach \x in {A, ..., Z}{%
	\expandafter\xdef\csname orcid\x\endcsname{\noexpand\href{https://orcid.org/\csname orcidauthor\x\endcsname}{\noexpand\orcidicon}}
}

\begin{document}
\title{TransICD: Transformer Based Code-wise Attention Model for Explainable ICD Coding}
%
%
\author{Biplob Biswas\inst{1}\orcidA{}
\and
Thai-Hoang Pham\inst{1,2}\orcidB{}
\and
Ping Zhang\inst{1,2}\orcidC{}}
\authorrunning{B. Biswas et al.}
%
\institute{Department of Computer Science and Engineering, The Ohio State University, Columbus OH 43210, USA \and
Department of Biomedical Informatics, The Ohio State University, Columbus OH 43210, USA\\
\email{\{biswas.102, pham.375, zhang.10631\}@osu.edu}}

\maketitle              
\begin{abstract}
International Classification of Disease (ICD) coding procedure which refers to tagging medical notes with diagnosis codes has been shown to be effective and crucial to the billing system in medical sector. Currently, ICD codes are assigned to a clinical note manually which is likely to cause many errors. Moreover, training skilled coders also requires time and human resources. Therefore, automating the ICD code determination process is an important task. With the advancement of artificial intelligence theory and computational hardware, machine learning approach has emerged as a suitable solution to automate this process. In this project, we apply a transformer-based architecture to capture the interdependence among the tokens of a document and then use a code-wise attention mechanism to learn code-specific representations of the entire document. Finally, they are fed to separate dense layers for corresponding code prediction. Furthermore, to handle the imbalance in the code frequency of clinical datasets, we employ a label distribution aware margin (LDAM) loss function. The experimental results on the MIMIC-III dataset show that our proposed model outperforms other baselines by a significant margin. In particular, our best setting achieves a micro-AUC score of 0.923 compared to 0.868 of bidirectional recurrent neural networks. We also show that by using the code-wise attention mechanism, the model can provide more insights about its prediction, and thus it can support clinicians to make reliable decisions. Our code is available online\footnote{\url{https://github.com/biplob1ly/TransICD}}.

\keywords{ICD  \and Multi-label Classification \and Transformer-based Model}
\end{abstract}

\section{Introduction}
\begin{figure}
    \centering
    \includegraphics[width=0.8\textwidth]{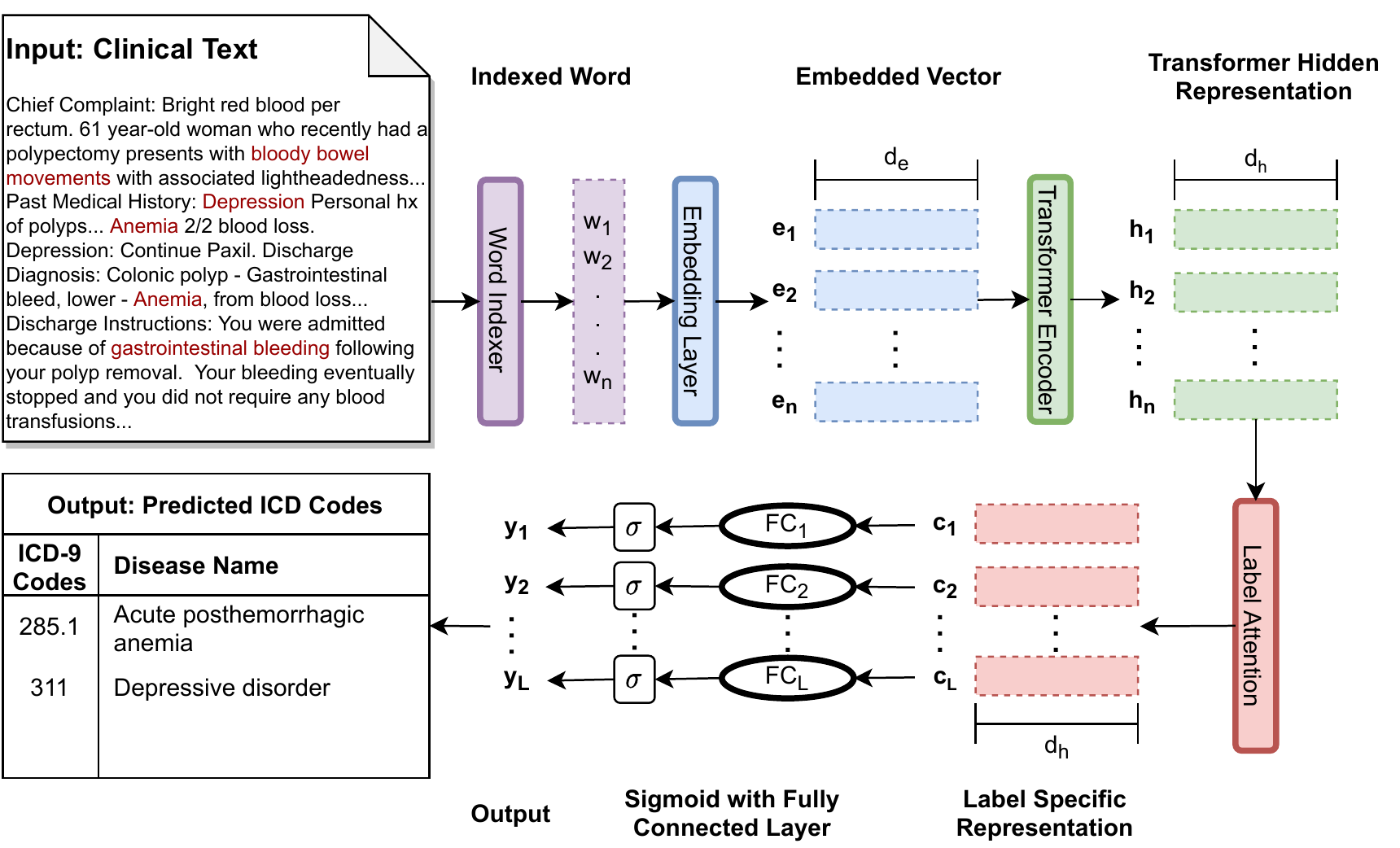}
    \caption{The framework of the proposed ICD coding model. The model takes clinical text as input and passes it through embedding layer, transformer encoding layer, label attention layer and finally through dense layer to predict corresponding codes.}
    \label{fig:icd_model}
\end{figure}

The International Classification of Diseases (ICD) is a health care classification system maintained by the World Health Organization (WHO)~\cite{who14}, that provides a unique code for each disease, symptom, sign and so on. Over 100 countries around the world use ICD codes and in the United States alone, the healthcare coding market is a billion-dollar industry~\cite{usmarket19}. In manual ICD coding, professional coders use patients’ clinical records representing diagnoses and procedures performed during patients’ visits to assign codes. While it serves purposes including billing, reimbursement and epidemiological studies, the task is expensive, time-consuming and error-prone. Fortunately, the advent of machine learning approaches has paved the way for automatic ICD coding. \autoref{fig:icd_model} illustrates an example of such ICD coding process where the coding model takes clinical text as input and outputs predicted ICD codes. It also shows that the model puts attention to subtext (highlighted in red) that is relevant to a disease, e.g. `gastrointestinal bleeding' is related to the disease \textit{Acute posthemorrhagic anemia' (ICD-9 code: 586)}. 

However, the task poses a couple of challenges. First, with more than 15,000 codes in ICD-9, it is a multi-label classification problem of high dimensional label space. Second, the majority of the codes are associated with rare diseases and hence, used infrequently, resulting in an imbalance in the dataset. Third, clinical records are noisy, lengthy and contain a large amount of medical vocabulary.

Previous well-known models~\cite{mullenbach18,shi2017} employed methods such as CNNs, LSTMs to automate ICD coding. However, CNNs and LSTMs have a weakness to encode the long sequence of discharge summaries (average token count before preprocessing $\approx1500$). On the other hand, a self-attention based transformer~\cite{vaswani2017} model processes a sequence as a whole and thus can avoid long term dependency issue of LSTMs. Unfortunately, most pre-trained transformer models such as off-the-shelf BERTs~\cite{devlin2018,alsentzer2019,lee2019} have a limitation of a smaller sequence length and the usual ones~\cite{devlin2018} experience a lot of out-of-vocabulary (OOV) words in representing clinical text. Training a transformer encoder with a pre-trained CBOW (Continuous Bag Of Words)~\cite{mikolov2013efficient} embedding of clinical tokens can mitigate both the problem of limited sequence length and OOV words.
With this intuition, in this work, we present an end-to-end deep-learning model for ICD coding. Here are our contributions:
\begin{itemize}
    \item We propose an ICD coding model that utilizes transformer encoder to obtain contextual representation of tokens in a clinical note. Aggregating those representations, we employ the structured self-attention mechanism~\cite{lin17} to extract label-specific hidden representations of an entire note.
    \item To address the long-tailed distribution of ICD codes, we apply a label distribution aware margin (LDAM)~\cite{cao19} loss function. For evaluation, we make a comparative analysis of our model with the well-known models on the benchmark MIMIC-III dataset~\cite{johnson16}.
    \item Finally, we present a case study to demonstrate visualizable attention to label-specific subtext indicating interpretability of our coding process.
\end{itemize}

\section{Related Works}
The study of automatic ICD coding can be traced back to the late 1990s~\cite{larkey96,lima98}. Last two decades have seen quite a good number of ICD coding models with various approaches from both feature-based classical machine learning and deep learning technique. Most of these studies addressed the task as a multi-label classification problem.

Larkey and Croft~\cite{larkey96} adopted an ensemble of K-nearest neighbors, relevance feedback and Bayesian independence to identify ICD code of a discharge summary. Both de Lima et al.~\cite{lima98} and Perotte et al.~\cite{perotte13} proposed hierarchical models to capture the hierarchical relationship of ICD codes. However, the former study uses cosine similarity between the discharge summaries while the latter one employs SVM for prediction.

In the last few years, different variations of neural networks have been applied to this task. Ayyar et al.~\cite{ayyar2016} and Shi et al.~\cite{shi2017} utilized word and character level LSTM (C-LSTM-Att) respectively to capture the long-distance relationships within a clinical text. Mullenbach~\cite{mullenbach18} employed the baseline models such as Logistic Regression (LR), CNN~\cite{kim2014}, Bi-GRU~\cite{cho2014} on the MIMIC datasets for ICD coding and presented a convolutional attention network (CAML) that achieved a state of the art results. 
In another work~\cite{Baumel2018}, the authors introduced a hierarchical attention as part of a GRU-architecture that provides interpretability. Wang et al. put forward a label embedding attentive model (LEAM)~\cite{wang2018} that encodes labels (i.e. codes) and words in the same representational space and uses cosine similarity between them for label prediction.
However, being motivated by the recent success of transformer-based models~\cite{devlin2018,lee2019,alsentzer2019}, in our ICD coding task, we train one such encoder from scratch to circumvent sequence length limitation and learn better token representation.

\section{Dataset}
MIMIC-III~\cite{johnson16} is one of the benchmark datasets that provides ICU medical records and is widely used in ICD coding prediction. Each record of it includes a discharge summary describing diagnoses and procedures that took place during a patient's stay and is labeled with a set of ICD-9 codes by professional coders. 
Following previous works~\cite{mullenbach18}, we prepare two common settings of the dataset: MIMIC-III full and MIMIC-III 50. In total, the MIMIC-III full setting contains 52,726 sets of discharge summaries and 8,929 unique codes. 6,918 of the codes are diagnosis codes and the rest 2,011 are procedure codes. Only 1.84\% of the diagnosis codes are assigned to more than 1000 discharge summaries, and the majority (87.5\%) of the ICD codes are tagged on to less than 100 notes, indicating an extremely long-tail of distribution.

Hence, we choose the MIMIC-III 50 setting which consists of the 50 most frequent ICD codes with 11,368 set of discharge summaries. The dataset is split into train, validation and test set by patient ID so that the test or validation set does not contain any patient data already seen in the training set. \autoref{tab:dataset} provides the summary of the dataset.

\begin{table}
\centering
\small
\def\arraystretch{1.5}
\addtolength{\tabcolsep}{3pt}
    \caption{The statistics for the data samples of the 50 most frequent ICD-9 codes in MIMIC-III dataset after preprocessing.}
    \label{tab:dataset}
    \begin{tabular}{l|c|c|c|c|c}
    \toprule
     \textbf{Split} & \textbf{\# Samples} & \makecell{\textbf{\# Unique}\\\textbf{Codes}} & \makecell{\textbf{\# Mean}\\\textbf{Tokens}} & \makecell{\textbf{\# Mean}\\\textbf{Codes}}  & \makecell{\textbf{\# Stdev of}\\\textbf{Code freq}}\\
    \toprule
    \rowcolor{lightgray}
    Train       & 8,066 & 50  &  922 &  5.69 & 577.89\\
    Validation & 1,573 & 50  &  1,115  &   5.88  &  121.01\\
    \rowcolor{lightgray}
    Test & 1,729 & 50  &  1,133   & 6.03  & 136.93\\
    \bottomrule
    \end{tabular}
\end{table} 

\textbf{Preprocessing.}\label{preprocess} For each discharge summary sample, we lowercase and tokenize the text, remove punctuations, numbers, English stopwords, and any token with less than three characters. After that, we stem them with Snowball stemmer and replace any remaining digits with character `n' which converts tokens such as `350mg' to `nnnmg'. 
From the resulting distribution of token count per record, we observe that more than 98\% of the discharge summaries are bound within 2500 tokens. So we use 2500 as the maximum length of token sequence for training. We exploit word2vec CBOW method~\cite{mikolov2013efficient} to obtain word embeddings of size, $d_e=128$ by training the entire discharge summary set. Finally, we extract a vocabulary of 123916 tokens from training set and augment it with `PAD' and `UNK' token for padding and out of vocabulary words respectively. 

\section{Methods}
\subsection{Problem Formulation}
Since each discharge summary sample can have multiple ICD codes associated with it, we approach the code prediction task as a multi-label classification problem. Given a clinical record with token sequence, $\textbf{W}=[w_1, w_2,..., w_n]$, our objective is to determine $y_{_{l \in L}} \in \{0, 1\}$ where $L$ is the set of labels i.e. ICD-9 codes. 

\subsection{Transformer Based Label Attention Model}
We leverage the concept of multi-headed self-attention, popularly known as transformer, to encode the tokens of the clinical notes. \autoref{fig:icd_model} illustrates the overall architecture of our model. 
The following subsections describe the model framework in detail.

\subsubsection{Embedding Layer}
Considering an input clinical note, $\textbf{W}=[w_1, w_2, \dots, w_n]^T$, where $w_i$ is the vocabulary index of the $i$-th word and $n$ is the maximum possible length, we map them to the pre-trained embeddings (\S\ref{preprocess}). This provides us with a matrix representation of the document, $\textbf{E}=[\textbf{e}_1, \textbf{e}_2, \dots, \textbf{e}_n]^T$ where $\textbf{e}_i \in \mathbb{R}^{d_e}$ is the word embedding vector for the $i$-th word.

\subsubsection{Transformer Encoder Layer}
The word embeddings, $\textbf{E} \in \mathbb{R}^{{n} \times {d_e}}$ of a clinical note is fed into a transformer encoder which employs multi-headed self attention mechanism~\cite{vaswani2017} to the sequence as a whole and provides us with contextual word representations, $\textbf{H} \in \mathbb{R}^{{n} \times {d_h}}$. Mathematically:
\begin{equation}
\textbf{H} = \transencoder(\textbf{E})
\label{eq:encoder}
\end{equation}
where $\textbf{H}=[\textbf{h}_1, \textbf{h}_2, \dots, \textbf{h}_n]^T$.

\subsubsection{Code-specific Attention Model}
Being a multi-label classification task, it demands further processing of the encoded representation, $\textbf{H} \in \mathbb{R}^{{n} \times {d_h}}$ to produce a code-wise representation. To this end, we apply a structured self-attention mechanism on $\textbf{H}$. First, the attention weights, $a_l \in \mathbb{R}^n$ corresponding to tokens of a note for label $l$ is computed by:
\begin{align}
& \textbf{a}_l = \softmax(\tanh(\textbf{H}\textbf{U})\textbf{v}_l)
\label{eq:attention_score}\\
& \textbf{c}_l = \textbf{H}^T{\textbf{a}_l}
\label{eq:attended_output}
\end{align}
where $\textbf{U} \in \mathbb{R}^{d_{h} \times d_{a}}$ and $\textbf{v}_l \in \mathbb{R}^{d_{a}}$ are trainable parameters and $d_{a}$ is a hyper parameter. Next, we multiply the contextual representation $\textbf{H}$ and the attention scores $\textbf{a}_l$ to produce a fixed length code-specific document representation $\textbf{c}_l$ for each label $l \in L$ (Eq. \ref{eq:attended_output}).
Intuitively, $\textbf{c}_l \in \mathbb{R}^{d_h}$ encodes information sensitive to label $l$. Finally, we concatenate this attended document representation $\textbf{c}_l$ for all labels to obtain $\textbf{C}=[\textbf{c}_1, \textbf{c}_2, \dots, \textbf{c}_L]^T \in \mathbb{R}^{{L} \times {d_h}}$

\subsubsection{Multi-label Classification}
To compute the probability for label $l$, we feed the corresponding label-wise document representation $\textbf{c}_l$ to a single layer fully connected network with a one node in the output layer followed by a sigmoid activation function (Eq. \ref{eq:output}). 
Having the probability score, We use a threshold of 0.5 to predict the binary output $\in \{0,1\}$.
For training, we adopt multi-label binary cross-entropy as loss function (Eq. \ref{eq:bceloss}).
\begin{align}
& \hat{y}_{l} = \sigma({\textbf{Z}\textbf{c}_l}+\textbf{b})
\label{eq:output}\\
& L_{BCE}(y, \hat{y}) = -\sum_{l=1}^{L}[y_l\log(\hat{y}_l) + (1-y_l)\log(1-\hat{y}_l)]
\label{eq:bceloss}
\end{align}
To address the long-tailed distribution of ICD codes in the dataset, following previous work of Song et al.~\cite{song20}, we employ label-distribution-aware margin (LDAM)~\cite{cao19}, where the probability score is computed by Eq.~\ref{eq:ldam_output}.
\begin{align}
& \hat{y}_{l}^{m} = \sigma({\textbf{Z}\textbf{c}_l}+\textbf{b} - \textbf{1}(y_l=1)\Delta_l)
\label{eq:ldam_output}\\
& L_{LDAM} = L_{BCE}(y, \hat{y}^m)
\label{eq:ldam_loss}
\end{align}
where function $\textbf{1}(.)$ is 1 if $y_l=1$. $\Delta_l=\frac{C}{n_l^{1/4}}$ and C is a constant and $n_l$ is the total count of training notes having $l$ as true label. Finally, we obtain LDAM loss using Eq.~\ref{eq:ldam_loss}.

\section{Training Details}
\label{sec:training_details}
A search for optimal hyper-parameter leads us to the following setting of values: \{Encoder layer: 2, Attention head: 8, Epochs: 30, Learning rate: 0.001, Dropout rate: 0.1\}. We also set $d_a=2*d_e$ and $C=3$. We train the models on an NVIDIA Tesla P100 (Pascal). In our best setting, each epoch takes around 168 seconds.

\section{Evaluation}
To evaluate our model, we utilize commonly used metrics such as micro-averaged and macro-averaged area under the ROC curve (AUC) and F1 score. As specified by Manning et al.~\cite{manning2008}, macro-averaged values are computed by averaging metrics calculated per label. On the other hand, micro-averaged values are computed considering each pair (document, code) as a separate prediction. The macro-averaged values are usually low in this task as they put more emphasis on infrequent label prediction. We also include precision at k (P@k) which computes the fraction of the true labels that are present in our top-k predictions. As the average number of codes per note is around 5.8, we choose $k=5$ for evaluation.

\subsection{Results}
\autoref{tab:results} provides a comparison of our proposed ICD coding model to the previous methods on the top-50 frequent ICD codes of the MIMIC-III dataset. The scores are in percentage and are measured on the held-out test set with the aforementioned hyperparameter setting (\S\ref{sec:training_details}). We ran our model five times and use different random seeds in each run to initialize the model parameters. We present the means and standard deviations of these five runs as our final result of the proposed TransICD model. The low standard deviations indicate that our model consistently performs well, and thus it is stable.

\begin{table}
\centering
\def\arraystretch{1.4}
\addtolength{\tabcolsep}{1.3pt}
    \caption{Test set results (in \%) of the proposed models on the MIMIC-III 50 dataset. Models marked with * are ours and values with boldface are the best in the corresponding column.}
    \label{tab:results}
    \begin{tabular}{l |c c|c c|c}
    \toprule
    \multirow{2}{*}{\textbf{Models}} & \multicolumn{2}{c|}{\textbf{AUC}} & \multicolumn{2}{c|}{\textbf{F1}} & \multirow{2}{*}{\textbf{P@5}}\\
        \cline{2-5}
      &  \textbf{Macro} & \textbf{Micro} & \textbf{Macro} & \textbf{Micro}\\
    \toprule
    \rowcolor{lightgray}
    Logistic Regression (LR) & 82.9 & 86.4 & 47.7 & 53.3 & 54.6\\
    Bi-GRU  & 82.8 & 86.8 & 48.4 & 54.9 & 59.1\\
    \rowcolor{lightgray}
    C-MemNN~\cite{prakash2017} & 83.3 & - & - & - & 42.0\\
    C-LSTM-Att~\cite{shi2017} & - & 90.0 & - & 53.2 & -\\
    \rowcolor{lightgray}
    CNN~\cite{kim2014} & 87.6 & 90.7 & \textbf{57.6} & 62.5 & \textbf{62.0}\\
    CAML~\cite{mullenbach18} & 87.5 & 90.9 & 53.2 & 61.4 & 60.9\\
    \rowcolor{lightgray}
    LEAM~\cite{wang2018} & 88.1 & 91.2 & 54.0 & 61.9 & 61.2\\
    \bottomrule
    *Transformer & 85.2 & 88.9 & 47.8 & 56.3 & 56.5\\
    \rowcolor{lightgray}
    *Transformer + Attention & 88.2 & 91.1 & 49.4 & 59.3 & 59.6\\
    \makecell[l]{*TransICD(Transformer + \\Attention + $L_{LDAM}$)} & $\textbf{89.4} \pm 0.1$ & $\textbf{92.3} \pm 0.1$ & $56.2 \pm 0.4$ & $\textbf{64.4} \pm 0.3$ & $61.7 \pm 0.3$\\
    \bottomrule
    \end{tabular}
\end{table} 

Our proposed TransICD model produced the highest scores on micro-F1, macro-AUC, and micro-AUC, whereas the result in macro-F1 and precision@5 are comparable to the corresponding best score. 
\autoref{tab:results} also shows that we achieved a substantial improvement from all the baselines including the recurrent networks (Bi-GRU, C-LSTM-Att) and convolutional models(CNN, CAML). 
In fact, our basic transformer model (without attention) that simply uses mean pooling over the encoded token vectors for document representation outperforms logistic regression (LR) by at least 2.3\% in macro-AUC, 2.5\% in micro-AUC, 0.1\% in macro-F1, 3.0\% in micro-F1, and 1.9\% in precision@5. We believe this is due to the transformer encoder's superior ability to capture the long-term dependency of the tokens in contrast to that of recurrent units or hand-crafted feature extraction. 

With code-wise attention and LDAM loss, our best setting TransICD exceeds the strong baseline LEAM~\cite{wang2018} in macro-AUC by 1.3\%, in micro-AUC by 1.1\%, in macro-F1 by 2.2\%, in micro-F1 by 2.5\% and in precision@5 by 0.5\%. In macro-F1, our model takes a back seat only to CNN~\cite{kim2014}. The overall low scores in this metric also signify that the models struggle in predicting rare codes. The precision@5 of our model indicates that out of 5 predictions with the top probabilities, on average 61.9\% i.e. 3.085 are correct. The score is relatively higher than most of the other baselines except CNN.

Comparing previous models, we observe that logistic regression (LR), being a conventional machine learning model, performs worse than all other neural networks. Further inspection reveals that the attention-based models result in a significant improvement over the normal ones of the same kind. For instance, CAML outperforms the regular CNN.

\paragraph{\textbf{Ablation Study}} 
The contribution of different components of our model can be recognized from the bottom three rows of \autoref{tab:results}.
First, we notice a substantial drop in every metric when label-distribution aware margin (LDAM) loss is not adopted. In another way, LDAM improves the performance in AUC by (macro-1.2\%, micro-1.2\%), F1 by (macro-6.8\%, micro-5.1\%), and precision@5 by 2.1\%. This clearly demonstrates that LDAM loss played a powerful role to counter the imbalanced frequency of the labels. Moreover, instead of the label attention, if we simply use mean pooling of the token representations from our transformer encoder to encode the entire document, we end up having the same hidden vector for all the labels. This further hurts the performance of the model. Putting differently, extending basic transformer model with code-wise attention increases AUC score by (macro-3.0\%, micro-2.2\%), F1 score by (macro-1.9\%, micro-3\%), and precision@5 by 3.1\%. This corroborates that extraction of code-specific representation of a document does improve the corresponding label prediction.

\subsection{Distribution of Scores}
\begin{figure}
    \centering
    \includegraphics[width=1.0\textwidth]{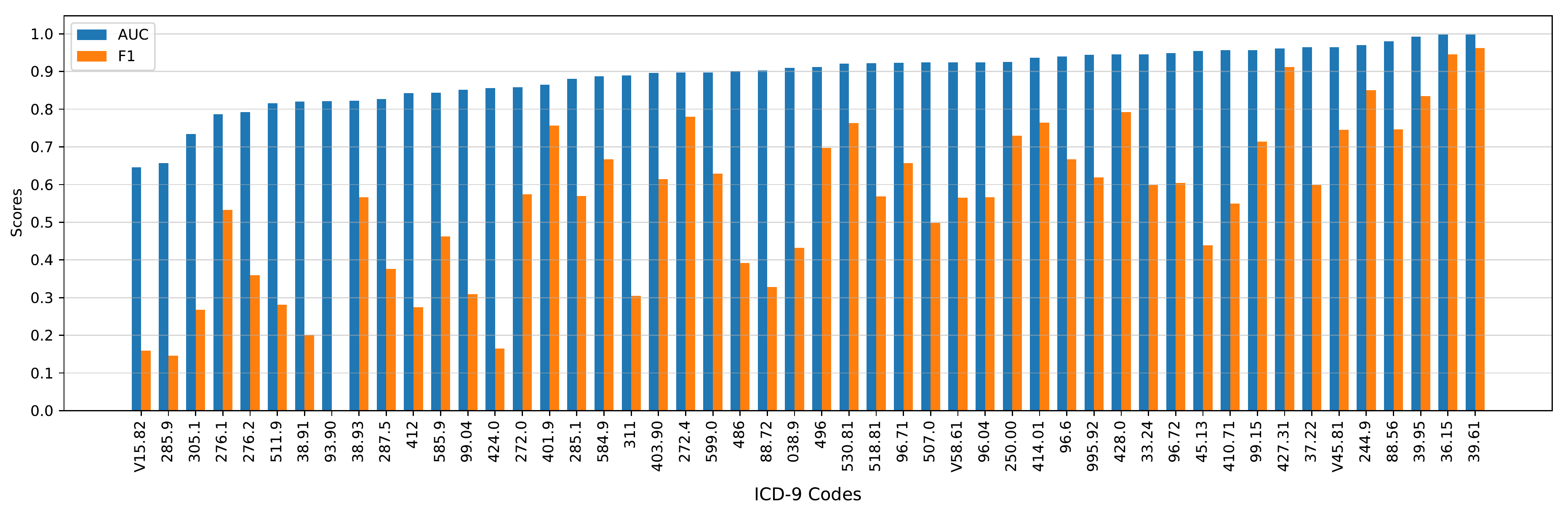}
    \caption{AUC and F1 scores across the top-50 frequent ICD-9 codes of MIMIC-III dataset}
    \label{fig:auc_f1}
\end{figure}
Our model achieves higher AUC scores for many ICD codes. Specifically, for 90\% of the codes, our model attains an AUC higher than 0.8 and for 56\% of them, we have an AUC higher than 0.9. On the other side, an AUC score lower than 0.7 is seen for only 4\% of the codes. We notice that some of the low scoring ICD codes such as V15.82, 305.1, 276.1 are also the least frequent ones in the training set.
Another observation shows misclassification among closely related codes. For instance, \textit{Tobacco use disorder (ICD: 305.1)} and \textit{Arterial catheterization (ICD: 38.91)} are seen to be very frequently mislabeled as \textit{History of tobacco use (ICD: V15.82)} and \textit{Venous catheterization, not elsewhere classified (ICD: 38.93)} respectively. Above all, most frequent wrongly classified codes such as 401.9, 96.04 are also found to be the dominant ones in the training set indicating a bias towards them. 
A naive random oversampling of the dataset can be a way to get rid of such bias. Analyzing F1 scores, we find a relatively smaller number of the codes (10\%) having F1 score greater than 0.8. We present the individual AUC and F1 score of the most frequent 50 codes in \autoref{fig:auc_f1}.

\subsection{Visualization}
\begin{figure}
    \centering
    \begin{tabular}{@{}c@{}c}
        (a) & \includegraphics[align=c, width=0.9\textwidth]{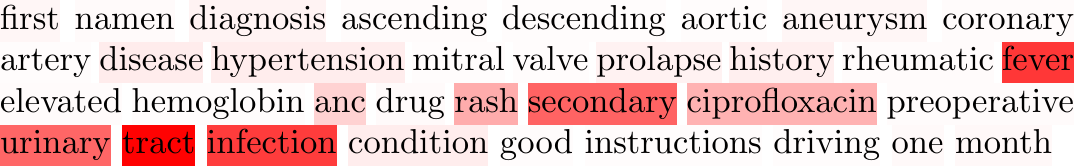}
        \label{fig:urinary}
  \end{tabular}
    \\
    \begin{tabular}{@{}c@{}}
        (b) \includegraphics[align=c, width=0.9\textwidth]{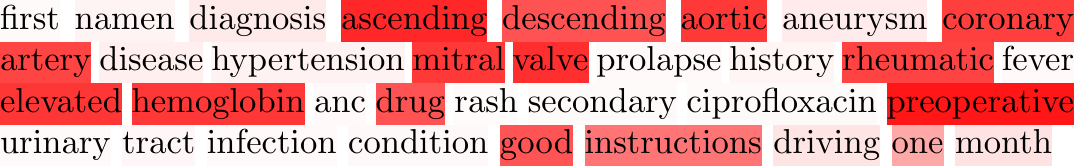}
        \label{fig:coronary}
      \end{tabular}
    \caption{Visualization of the model attending on an excerpt from a discharge summary for label- (a) \textit{Urinary tract infection (ICD: 599.0)} and (b) \textit{Single internal mammary-coronary artery bypass (ICD: 36.15)}. Darker color indicates higher attention.}%
    \label{fig:visualization}%
\end{figure}

For high-stakes prediction applications such as healthcare, there has been an increasing demand to explain the prediction of a model in a way that humans can understand. Although an automated model is set to reduce human labor, being able to observe which parts of a text are contributing to the final prediction provides reliability and transparency. In \autoref{fig:visualization}, we provide such visualization of our code-wise attention model where an excerpt of a note is highlighted with attention scores corresponding to two different labels.

Figure~\autoref{fig:urinary} shows that for disease \textit{Urinary tract infection (ICD: 599.0)}, our model successfully puts high attention to the closely related words- `urinary tract infection'. However, the model ignores the same words while predicting for label \textit{Single internal mammary-coronary artery bypass (ICD: 36.15)} as illustrated in Figure~\autoref{fig:coronary} because they are not relevant for the latter label. On the other hand, being associated with the latter label, `coronary artery' is seen to gain more attention in Figure~\autoref{fig:coronary}, although the same bi-gram is not attended for the former label in Figure~\autoref{fig:urinary}.

All these suggest that the reasoning of our model is highly correlated to the features that a human would have looked for while tagging a note with ICD codes. Consequently, we believe, this model would help clinicians in the ICD coding process with higher reliability and transparency.

\section{Conclusion}
The study proposes a transformer-based deep learning method to predict ICD codes from discharge summaries representing diagnoses and procedures conducted during patients' stay in hospital. We adopt LDAM loss to counter the imbalanced dataset and employ a code-wise attention mechanism for more accurate multi-label predictions. Our visualization report illustrates that the model attends to the relevant features and hence provides evidence for reliability. For future work, we will focus on a larger dataset containing more or even all the ICD codes.

%
%


\end{document}